\documentclass[conference]{IEEEtran}
\IEEEoverridecommandlockouts
\usepackage{cite}
\usepackage[numbers]{natbib}
\usepackage{amsmath,amssymb,amsfonts,bm}
\usepackage{algorithmic}
\usepackage{algorithm}
\usepackage{graphicx}
\usepackage{subcaption}
\usepackage{textcomp}
\usepackage{booktabs} 
\DeclareMathOperator*{\argmin}{arg\,min}

\usepackage{pgf}
\usepackage{tikz}
\usetikzlibrary{arrows,automata}
\usepackage[latin1]{inputenc}
\usepackage{verbatim}

\def\code#1{\texttt{#1}}

\def\BibTeX{{\rm B\kern-.05em{\sc i\kern-.025em b}\kern-.08em
    T\kern-.1667em\lower.7ex\hbox{E}\kern-.125emX}}
\begin{document}

\title{VOS: a Method for Variational Oversampling of Imbalanced Data}

\author{\IEEEauthorblockN{Val Andrei Fajardo}
\IEEEauthorblockA{\textit{integrate.ai} \\
Toronto, Canada \\
andrei@integrate.ai}
\and
\IEEEauthorblockN{David Findlay}
\IEEEauthorblockA{\textit{integrate.ai} \\
Toronto, Canada \\
david@integrate.ai}
\and
\IEEEauthorblockN{Roshanak Houmanfar}
\IEEEauthorblockA{\textit{integrate.ai} \\
Toronto, Canada \\
roshan@integrate.ai}
\and
\IEEEauthorblockN{Charu Jaiswal}
\IEEEauthorblockA{\textit{integrate.ai} \\
Toronto, Canada \\
charu@integrate.ai}
\and
\IEEEauthorblockN{Jiaxi Liang}
\IEEEauthorblockA{\textit{integrate.ai} \\
Toronto, Canada \\
garcia@integrate.ai}
\and
\IEEEauthorblockN{Honglei Xie}
\IEEEauthorblockA{\textit{integrate.ai} \\
Toronto, Canada \\
holly@integrate.ai}
}

\maketitle
\begin{abstract}
Class imbalanced datasets are common in real-world applications that range from credit card fraud detection to rare disease diagnostics. Several popular classification algorithms assume that classes are approximately balanced, and hence build the accompanying objective function to maximize an overall accuracy rate. In these situations, optimizing the overall accuracy will lead to highly skewed predictions towards the majority class. Moreover, the negative business impact resulting from false positives (positive samples incorrectly classified as negative) can be detrimental. Many methods have been proposed to address the class imbalance problem, including methods such as over-sampling, under-sampling and cost-sensitive methods. In this paper, we consider the over-sampling method, where the aim is to augment the original dataset with synthetically created observations of the minority classes. In particular, inspired by the recent advances in generative modelling techniques (e.g., Variational Inference and Generative Adversarial Networks), we introduce a new oversampling technique based on variational autoencoders. Our experiments show that the new method is superior in augmenting datasets for downstream classification tasks when compared to traditional oversampling methods.
\end{abstract}


\section{Introduction}
Imbalanced datasets pose a problem in machine learning classification tasks and are present in a multitude of real-world industry datasets. These datasets are characterized as having class priors that are vastly different from one another, and that are skewed towards a majority class or classes (e.g. see Liu and Ghosh \citep{liu2007generative}). In the case of a dataset with binary classes, an imbalance would cause the minority class to have a significantly smaller prior than the majority class, whereas a balanced dataset would have similar class priors.

Imbalanced datasets occur in a variety of industries such as retail banking, insurance, and telecommunications, with applications in fraud detection, customer acquisition, etcetera. In these cases, it is most critical to correctly classify the minority class. Incorrectly labelling a sample as a false positive can have harsh consequences and business risks, for example in the case of incorrectly labelling a credit card transaction as fraudulent and unnecessarily penalizing a customer for this transaction.

The challenge with imbalanced datasets arises as a result of classifier bias towards majority class predictions, given that their objective function does not consider class differences \citep{maloof2003learning}. The canonical methods of addressing class imbalances include sampling techniques \citep{chawla2002smote}, changing cost functions \citep{mccarthy2005does}, and algorithm level methods. 

In SMOTE \cite{chawla2002smote}, the minority class is over-sampled by creating synthetic examples in the same feature space as the data. The synthetic examples lie on the line segments that join K minority class neighbours. This technique forces the decision region of the minority class to be more general. Altering the class distributions of a dataset does have downsides, however; under-sampling the majority class may lead to discarding useful data and oversampling the minority class can lead to overfitting \citep{mccarthy2005does}. 

With ADASYN \citep{he2008adasyn}, the authors adaptively generate examples of the minority class, according to the distributions of the minority samples. More synthetic data is generated for minority samples that are difficult to learn, versus those that are easier to learn. The algorithm uses a density distribution to determine the number of additional synthetic examples needed to be generated for each minority sample. This is in contrast to SMOTE, where an equal amount of synthetic data are generated for each minority data sample. 

In contrast to SMOTE and ADASYN, cost-sensitive learning techniques do not modify the imbalanced data distribution directly  \citep{elkan2001foundations}. Instead the problem is targeted by using different cost-matrices to describe the cost of misclassifying data samples as false negatives or false positives.  These techniques can also consider learning when error costs are unequal \citep{maloof2003learning}. When misclassification costs are known, they can be incorporated directly into the cost function \citep{mccarthy2005does}. In \citep{mccarthy2005does}, the authors showed that cost-sensitive learning and oversampling perform similarly with no definitive winner between cost-sensitive, undersampling, or oversampling. 

In contrast, generative approaches have shown promise by outperforming traditional sampling or cost-sensitive techniques \citep{liu2007generative}. The authors generated synthetic data points from the minority class by first learning the probability distribution of the minority class and subsequently adding to a resampled set until the desired proportion between minority and majority classes was reached. They generated artificial documents by sampling from the learned multinomial distribution of the minority class with the objective of applying these documents for word prediction. 

In this paper, we similarly focus on generative methods for oversampling and introduce a new generative modelling approach using Variational Autoencoders (VAE) to oversample the minority class in an imbalanced dataset, with a focus on binary target variables. However, the approach can be used easily in multi-class situations. We also extend our approach to image datasets, and allow our architecture to work with convolutional neural networks (CNN). Furthermore, to the best of our knowledge, the research conducted in this paper is the first of its kind to apply variational inference to oversample minority classes when dealing with imbalanced datasets. 

The remainder of the paper is organized as follows. In the next section, we briefly review the Variational Autoencoder. In Section 3, we introduce the new generative model for computing synthetic observations of the minority class. The results of an application of the new method to a large real-world dataset is discussed in Section 4, where we show that the new method outperforms SMOTE with respect to a downstream binary classification task. Finally, in Section 5 we conclude with some closing remarks on the new method and present potential avenues for future research. 

\section{Variational Autoencoders}
Variational methods are employed in situations where the computation of complex integrals are not feasible (i.e., due to either mathematical intractability or extreme computational complexity). The essential idea in variational methods is to approximate the integrand, say $f(x)$, with a more simple to integrate function, say $q(x)$, and allow the algorithm to improve $q(x)$ based on some a priori distributions. Variational Autoencoders were first introduced by \citet*{kingma2013auto}. With VAEs, we are able to perform efficient approximate inference when learning probabilistic models whose (continuous) latent variables have intractable posterior distributions. Moreover, the objective function for VAEs is formed by obtaining a lower bound to the log marginal likelihood of the data, which is typical when learning latent variable models with variational inference. This function is specifically called the evidence lower bound (ELBO), and is given by
\begin{multline}
\log p_\theta(x) \geq E_{q_\phi(z|x)}[\log p_\theta(x | z)] \\
- KL(q_\phi(z|x) || p_\theta(z)),
\end{multline}
where $z$ is the latent variable, $q_\phi(z|x)$ is the variational distribution, and $KL(q||p)$ denotes the Kullback-Liebler divergence between two distributions $q$ and $p$. We get an unbiased estimate of ELBO by sampling $z$ and performing stochastic gradient ascent to optimize this \cite{liang2018variational} with respect to $phi$ and $\theta$. It should be note that in order to utilize back propagation, a reparametrization trick is applied in order to sample $z$. That is, we sample random noise $\epsilon$ and obtain $z = g(\epsilon)$, where $g(\cdot)$ is a continuous and differentiable function with respect to $\theta$ and $\phi$. Finally, the VAE is considered as a generative model, since it learns the conditional distribution $p_\theta(x|z)$. In other words, to sample $x$ from this distribution, one first randomly samples $z$ and then samples an observation of $x$ from the distribution of $p_\theta(x|z)$.

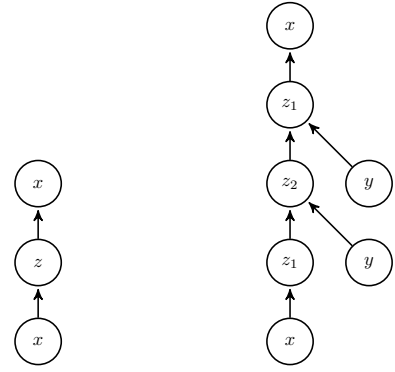
\begin{figure}[htbp]
\centering
\begin{subfigure}[b]{0.2\textwidth}
\centering
\begin{tikzpicture}[->,>=stealth',shorten >=1pt,auto,node distance=1.5cm,
                    semithick, scale=0.70]
\tikzstyle{every state}=[fill=white,draw=#1,text=black,scale=0.70]

\node[state]	(x) {$x$};
\node[state] (z) [above of=x] {$z$};	 
\node[state] (x-d) [above of=z] {$x$};	 

\path (x) edge  (z);
\path (z) edge  (x-d);

\end{tikzpicture}
\caption{Traditional VAE}
\end{subfigure}~
\begin{subfigure}[b]{0.2\textwidth}
\centering
\begin{tikzpicture}[->,>=stealth',shorten >=1pt,auto,node distance=1.5cm,
                    semithick, scale=0.70]
\tikzstyle{every state}=[fill=white,draw=#1,text=black, scale=0.70]

\node[state]	(x) {$x$};
\node[state] (z1) [above of=x] {$z_1$};	 
\node[state] (z2) [above of=z1] {$z_2$};
\node[state] (y) [right of=z1] {$y$};
\node[state] (z1-d) [above of=z2] {$z_1$};
\node[state] (y-d) [right of=z2] {$y$};
\node[state] (x-d) [above of=z1-d] {$x$};

\path (x) edge  (z1);
\path (z1) edge  (z2);
\path (y) edge  (z2);
\path (z2) edge  (z1-d);
\path (y-d) edge (z1-d);
\path (z1-d) edge (x-d);
\end{tikzpicture}
\caption{2-stage latent structure}
\end{subfigure}
\caption{Comparison of latent structures}
\label{vae-structures}
\end{figure}

\section{VOS: Variational Oversampling}

A VAE is comprised of two neural networks, one which learns the variational distribution $q_\phi (z|x)$, and another that learns the posterior distribution $p_\theta(x|z)$. Extensions of VAEs include those which consider several layers of latent variables, each layer requiring two neural networks, one for encoding and the other for decoding as described in the previous statement. 

The new approach is simple, and is one that requires only two stages of the latent structure: the first latent variable, $z_1$, encodes a pattern $x$, where as the second encoding $z_2$ can be seen as summarizing both the information of $z_1$ and the target label $y$. This approach was inspired by \citet*{louizos2015variational}, where the authors considered a two-stage latent structure to extract the features from a dataset, while removing the undesirable effect of sensitive features. We refer to this new oversampling method as VOS, which stands for Variational Oversampling.

The modified ELBO for the new VOS is derived in a similar manner as to that for the supervised case of the VFAE in \cite{louizos2015variational}. First, note that 
\begin{multline*}
p_\theta(x) =\int_{z_{1,n}}\int_{z_{2,n}}\int_{y_n} p_\theta(z_{2,n},y_n)p_\theta(z_{1,n}|z_{2,n},y)\\
\times p_\theta(x|z_{1,n})\mathrm{d}y_n\mathrm{d}z_{2,n}\mathrm{d}z_{1,n}. 
\end{multline*}
It then follows, after an application of Jensen's inequality that:
\begin{multline}\label{step1}
\sum_{i=1}^N \log p_{\theta}(x_i) \\ \geq  \sum_{i=1}^N E_{q_\phi(z_{1,n},z_{2,n},y_n|x_n)}\bigg[ \log(p_\theta(y_n) + \log p_{\theta}(z_{2_n})\\
 + \log p_{\theta}(z_{1,n}|z_{2,n},y_n) + \log p_\theta(x|z_{1,n})\\
- \log q_\phi(z_{1,n},z_{2,n},y_n|x_n)\bigg].
\end{multline}
Next, we assume that $q_\phi(x_n, z_{2,n}, y_n|z_{1,n}) = q_\phi(x_n|z_{1,n}) q_\phi(z_{2,n},y_n|z_{1,n})$, which then leads to the fact that 
\begin{multline}\label{step2}
q_\phi(z_{1,n},z_{2,n},y_n | x_n) = q_\phi(z_{1,n}|x_n)q_\phi(y_n|z_{1,n})\\
\times q_\phi(z_{2,n}|y_n,z_{1,n}).
\end{multline}

Finally, by combining Equations \eqref{step1} and \eqref{step2} we achieve the final desired loss function, namely:
\begin{multline}
\mathcal{L} = \sum_{n=1}^N\bigg( E_{q_\phi(z_{1,n}|x_n)}\big[\log p_\theta(x_n | z_{1,n})\\
- KL(q_\phi(z_{2,n}|z_{1,n},y_n) || p(z_{2,n}))\big] \\
+ E_{q_\phi(z_{2,n}|z_{1,n},y_n)}\big[-KL(q_\phi(z_{1,n}|x_n)||p_\theta(z_{1,n}|z_{2,n},y_n)\big]\bigg).
\end{multline}

The assumed parametric forms of the involved distributions are as follows:
\begin{align*}
q_\phi(z_{1,n}|x_n) &= \mathcal{N}(z_{1,n}|\mathbf{\mu}_n = f_\phi(x_n), \mathbf{\sigma}_n = e^{f_\phi(x_n)})\\
q_\phi(z_{2,n}|z_{1,n},y_n) &= \mathcal{N}(z_{2,n}|\mathbf{\mu}_n = f_\phi(z_{1,n},y_n),\\ &\qquad\qquad\mathbf{\sigma}_n = e^{f_\phi(z_{1,n},y_n)})\\
p_\theta(z_{1,n}|z_{2,n},y_n) &= \mathcal{N}(z_{1,n}|\mathbf{\mu}_n = f_\theta(z_{2,n},y_n),\\ &\qquad\qquad\mathbf{\sigma}_n = e^{f_\theta(z_{2,n},y_n)})\\
p_\theta(x_n|z_{1,n}) &\sim B(\Lambda = f_\theta(z_{1,n}))
\end{align*}
Note that $B(\Lambda)$ is an appropriate distribution whose parameters are denoted by $\Lambda$. For continuous variables, we assume that $B\sim \mathcal{N}(x_n|\mathbf{\mu}_n = f_\theta(z_{1,n}), \mathbf{\sigma}_n = e^{f_\theta(z_{1,n})})$; whereas for binary variables, we assume $B\sim Bernoulli(x_n|\mathbf{\rho}_n = f_\theta(z_{1,n}))$, where $\rho$ represents the probability that the random variable takes on the value of 1. It is also worth mentioning that all of the Gaussians above are assumed to have covariance structures whose off-diagonal elements are all zero (i.e., the dimensions of the latent representations are normallly distributed and independent of one another). 

\begin{algorithm}[htbp]
\caption{VAE generative model pseudo code}\label{pcode-vae}
\begin{algorithmic}[1]
\STATE $i \gets 1$
\WHILE {!convergence}
\STATE \%\% Encode X
\STATE $Z1_{mean}, Z1_{gamma} \gets \mathit{MLP}(X)$
\STATE $Z1 \gets \exp(0.5*Z1_{gamma})\times \mathit{noise}() + Z1_{mean}$
\STATE \%\% Encode Z1
\STATE $Z2_{mean}, Z2_{gamma} \gets \mathit{MLP}(Z1,Y)$
\STATE $Z2 \gets \exp(0.5*Z2_{gamma})\times \mathit{noise}() + Z2_{mean}$
\STATE \%\% Decode Z2
\STATE $Z1_{mean}, Z1_{gamma} \gets \mathit{MLP}(Z2,Y)$
\STATE $Z1 \gets \exp(0.5*Z1_{gamma})\times \mathit{noise}() + Z1_{mean}$
\STATE \%\% Decode Z1
\STATE $X_{mean}, X_{gamma} \gets \mathit{MLP}(Z1)$
\STATE $X \gets \exp(0.5*X_{gamma})\times \mathit{noise}() + X_{mean}$
\STATE \%\% Compute weight updates
\STATE $\phi \gets \phi + \alpha \frac{\mathrm{d}\mathcal{L}}{\mathrm{d}\phi}$
\STATE $\theta \gets \theta + \alpha \frac{\mathrm{d}\mathcal{L}}{\mathrm{d}\phi}$
\STATE $i \gets i + 1$
\ENDWHILE
\end{algorithmic}
\end{algorithm}

\section{Variational Methods for Image Data}

Previously documented methods of oversampling image data have included techniques like SMOTE and warping  (e.g., see \citet*{wong2016understanding}). In  \citep{wong2016understanding} , the authors defined a normalized random displacement field, such that each pixel in an image would be displaced by this vector. The displacement was governed by two parameters, $\alpha$ and $\sigma$, which controlled the strength and smoothness of the displacement. Testing their method on the MNIST dataset, however, they found that large displacements would result in images that no longer corresponded to the desired label. 

\citet*{zhang2015learning} created synthetic images of building roofs to augment their original dataset, but found a "synthetic gap" in the distributions of  the artificially generated images and the real images. The authors tried to train a sparse autoencoder simultaneously with real and synthetic images to minimize the synthetic gap. 

In \citep{pu2016variational}, the authors employ a new VAE-based method for deep deconvolutional learning, where a CNN is used in the encoder (as the recognition model) for the posterior distribution of the decoder, which functions as the image generative model. 

\section{Experiments}

In this section, we consider two separate imbalanced datasets and apply the new VOS method to oversample the minority class. In order to assess the performance of the oversampling technique, we train a classifier on the balanced dataset and record the performance on an untouched (i.e., unbalanced) test set. The accuracy metrics we use to judge the quality of oversampling (and classifier) are related to the receiver operating characteristic (ROC) graphs  \citep{fawcett2004roc}. Under imbalanced conditions, traditional overall accuracy would not provide a comprehensive view of the learning algorithm's performance  \citep{provostanalysis}. In particular, the metrics used to analyze the three conditions were F1-score, precision, and recall. 

It also bears mentioning that in both of our examples, we performed $K$-fold cross-validation to determine the number of hidden units in the hidden layers of the generative VAE. In particular, if we denote $L_i(a)$ as the final loss on the $i$-th heldout set when using architecture $a$; the the optimal architecture is given by
\begin{align}
a_{opt} = \argmin_{a} \frac{1}{K}\sum_{i=1}^K \mathcal{L}_i (a).
\end{align}

We use the \code{scikit-learn} implementation of logistic regression and set the accompanying parameters to their defaults, except for the inverse of regularization strength which was set to 10. For SMOTE as well ADASYN, we used the \code{imblearn} implementation with its the default parameters. Our experiments were run with four NVIDIA GRID GPUS, each with 1536 CUDA cores, 32 vCPUs, 60 GiB of memory, and 240 GB of SSD storage (i.e., using an AWS g2.8xlarge instance). Our implementation of VAEs is based in \code{Tensorflow}.

\subsection{Dataset 1: Credit Card Fraud Detection }

The credit card fraud detection dataset (e.g., see \cite{dal2015calibrating}) contains the transactions carried out by European cardholders over a two-day period in September 2013. Fraudulent transactions (i.e., the positive class) only accounted for 0.171\% of the total 341,762 transactions; and so, the dataset is highly imbalanced. We randomly split the set of transactions into a training set of 284,807 observations (492 of which were fraudulent), and a test set of the remaining 56,955 observations (91 of which were fraudulent).

For confidentiality purposes, the authors of the dataset were not able to provide the original features of the dataset,  hence they applied a PCA transformation to the original data to result in the obfuscated features that we used, which were essentially principal components. The untransformed features that were provided were time of transaction, transaction amount, class label. In total there were 31 features and the data itself only contained numerical variables. 

We note that for the cross-validation procedure to determine the architecture of the generative model, we set $K=5$ and restrict architectures to having a certain symmetric structure. This resulted in an optimal architecture wherein the hidden layers of the encoding and decoding layers in an consisted of 80 units, while both $z_1$ and $z_2$ to be of dimension 20. 

For the downstream classification task, we compare the results of three different classification algorithms, namely: logistic regression (LR), random forest (RF), and multi-layer perceptron (MLP). Furthermore, we also compare the accuracy metrics of the downstream task when trained on the resulting balanced datasets via SMOTE and ADASYN. We report the accuracy metrics for all pairs of oversampling techniques and classifiers in Table \ref{acc-metrics} (note that the predicted column represents the number of predictions of fraud transactions). 

\begin{table}[htbp]
\centering
\begin{tabular}{lrrrrr}
\toprule
method &  accuracy &  precision &  recall &  F1-score &  predicted \\
\midrule
          LR &     0.972 &      0.056 &   0.959 &    0.105 &  1689 \\
    SMOTE+LR &     0.972 &      0.056 &   0.959 &    0.105 &  1693 \\
   ADASYN+LR &     0.907 &      0.018 &   0.969 &    0.035 &  5404 \\
      VOS+LR &     0.999 &      0.802 &   0.908 &    0.852 &   111 \\
      & && & & \\
         MLP &     0.978 &      0.067 &   0.888 &    0.124 &  1304 \\
   SMOTE+MLP &     0.990 &      0.136 &   0.918 &    0.236 &   664 \\
  ADASYN+MLP &     0.989 &      0.131 &   0.949 &    0.230 &   709 \\
     VOS+MLP &     1.000 &      0.830 &   0.898 &    0.863 &   106 \\
     & && & & \\
          RF &     1.000 &      0.888 &   0.888 &    0.888 &    98\\
    SMOTE+RF &     1.000 &      0.905 &   0.878 &    0.891 &    95\\
   ADASYN+RF &     1.000 &      0.878 &   0.878 &    0.878 &    98\\
      VOS+RF &     0.999 &      0.667 &   0.918 &    0.773 &   135\\
\bottomrule
\end{tabular}
\caption{Accuracy metrics for fraud detection}
\label{acc-metrics}
\end{table}

As evidenced in Table \ref{acc-metrics}, oversampling with VAE significantly outperformed SMOTE as well as ADASYN and helped the classifier to achieve outstanding accuracy metrics on the test set. In particular, when using LR an MLP, the precision and F1 scores of the VAE were significantly higher than other two oversampling techniques; in addition, the overall accuracy is also higher. It is important to note that the performance of the RF without any oversampling techniques is comparable to SMOTE and ADASYN, while is much better than RF combined with VOS. The results of this experiment show the potential in applying variational inference for oversampling the minority class. We also note that in the \code{scikit-learn} implementation of both LR and MLP, that the \code{sample\_weight} parameter of the \code{fit} method enables one to weight synthetic observations differently from real ones. However, in our experiments, changing this value had no real significant impact. We set \code{sample\_weight} to 0.2 for synthetic observations on the basis that the predictive unit should not learn too much on the generated patterns relative to the real ones (i.e., \code{sample\_weight} was set to 1 for real observations).

\subsection{Dataset 2:  Tumour Images}

The second dataset used was the Breast Cancer Histopathalogical Image Classification (BreakHis) database  \citep{spanhol2016dataset}, which has 9109 microscope images of breast tumour tissue collected from 82 patients using a range of magnifications (40X, 100X, 200X, and 400X). It contains 2480 benign and 5429 malignant samples. This is an example of a use case where the cost of misclassification is very grave. Benign tumours are slow growing and localized, whereas malignant tumours are cancerous and can spread to other parts of the body to cause death. The training set had 4931 malign samples, and 2241 benign, whereas the test set had 498 malign and 239 benign samples. The same cross-validation procedure mentioned above was used on this image dataset as well. 

The png images of the breast cancer tumours were initially sized at 64 pixels x 64 pixels x 3 RGB colour channels. We flattened the images by turning each into a vector of dimension 64x64x3, and then applied a standard scalar across all of the images for normalization. The VOS algorithm was then used to oversample from this flattened vectors. Once oversampled, we reshape the flattened vectors into their original three dimensional shapes, for passing to the CNN classifier. We used three convolutional layers, with kernel size of 3x3 pixels, stride of 1, and 128 filters. We used ReLU activation functions, and applied dropout at each layer with a keep probability of 0.25. Max pooling operations were also used after each convolutional layer, and the last two layers of the network were fully connected with 1024 hidden units.

\begin{table}[htbp]
\centering
\begin{tabular}{lrr}
\toprule
method &  accuracy &     F1-score  \\
\midrule
          CNN &     0.900 &      0.926  \\
    VOS+CNN &     0.943 &      0.965 \\
\bottomrule
\end{tabular}
\caption{Accuracy metrics for fraud detection}
\label{image-acc-metrics}
\end{table}

We can see from Table \ref{image-acc-metrics}, that the new VOS methods also helps to improve the accuracy when compared to using CNN without balancing.

\section{Conclusion}
In this paper, we introduced a new generative approach for oversampling based on variational inference. In particular, we used a two-stage latent structure VAE to learn a sampling distribution of the original dataset. In order to learn the minority class distribution, the target responses augment $z_1$ encodings to learn the second encodings $z_2$. Our experimental results illustrated the superior performance of the new oversampling method versus SMOTE as well as ADASYN, and indeed demonstrate the promise of this new method for dealing with imbalanced datasets. 

With respect to future work, the authors are interested in testing variations of VAEs that lead to lower loss and thus better reconstructions such as Importance Weighted Autoencoders \cite{burda2015importance}. Learning richer covariance structures for the assumed Gaussian (i.e., relaxing the assumption of independence of the dimensions of the latent encodings $z_1$ and $z_2$) are also of interest, which we believe could also lead to lower reconstruction losses, and thereby more useful synthetic observations of the minority classes.

\bibliographystyle{IEEEtranN}
\bibliography{references}

\end{document}